\documentclass[lettersize,journal]{IEEEtran}
\usepackage{amsmath,amsfonts}
\usepackage{algorithmic}
\usepackage{algorithm}
\usepackage{array}
\usepackage[caption=false,font=normalsize,labelfont=sf,textfont=sf]{subfig}
\usepackage{textcomp}
\usepackage{stfloats}
\usepackage{booktabs}
\usepackage{url}
\usepackage{multirow}
\usepackage{verbatim}
\usepackage{graphicx}
\usepackage{cite}
\usepackage{hyperref}
\usepackage{orcidlink}
\usepackage{mcite}
\usepackage{tikz}
\hypersetup{
colorlinks=true,
linkcolor=blue,
citecolor=blue
}
\begin{document}

\title{Efficient Driving Behavior Narration and Reasoning on Edge Device Using Large Language Models}

\author{Yizhou Huang 
\hspace{-1.5mm}$^{~\orcidlink{0009-0008-2189-2470}}$,
Yihua Cheng \hspace{-1.5mm}$^{~\orcidlink{0000-0003-1353-9817}}$,
Kezhi Wang,~\IEEEmembership{Senior Member,~IEEE}
\hspace{-1.5mm}$^{~\orcidlink{0000-0001-8602-0800}}$

\thanks{This work was partly supported by Eureka i2D-MSW: intelligence to Drive | Move-Save-Win (Innovate UK project under Grant No. 10071278) and Horizon Europe COVER project, No. 101086228 (with funding from UKRI grant EP/Y028031/1). Kezhi would like to acknowledge the support in part by the Royal Society Industry Fellowship (IF/R2/23200104).}
\thanks{Yizhou Huang and Kezhi Wang are with the Department of Computer Science,  Brunel University London, UB8 3PH, UK (email: yizhou.huang2@brunel.ac.uk; kezhi.wang@brunel.ac.uk).}
\thanks{Yihua Cheng is with the School of Computer Science, University of Birmingham, B15 2TT, UK (email: y.cheng.2@bham.ac.uk)
}}

\markboth{Journal of \LaTeX\ Class Files,~Vol.~xx, xx.~8, xx~xxxx}%
{Shell \MakeLowercase{\textit{et al.}}: A Sample Article Using IEEEtran.cls for IEEE Journals}

\IEEEpubid{0000--0000/00\$00.00~\copyright~2021 IEEE}

\maketitle

\begin{abstract}
Deep learning architectures with powerful reasoning capabilities have driven significant advancements in autonomous driving technology. Large language models (LLMs) applied in this field can describe driving scenes and behaviors with a level of accuracy similar to human perception, particularly in visual tasks. Meanwhile, the rapid development of edge computing, with its advantage of proximity to data sources, has made edge devices increasingly important in autonomous driving. Edge devices process data locally, reducing transmission delays and bandwidth usage, and achieving faster response times. 
In this work, we propose a driving behavior narration and reasoning framework that applies LLMs to edge devices. The framework consists of multiple roadside units, with LLMs deployed on each unit. These roadside units collect road data and communicate via 5G NSR/NR networks. Our experiments show that LLMs deployed on edge devices can achieve satisfactory response speeds.
Additionally, we propose a prompt strategy to enhance the narration and reasoning performance of the system. This strategy integrates multi-modal information, including environmental, agent, and motion data. Experiments conducted on the OpenDV-Youtube dataset demonstrate that our approach significantly improves performance across both tasks.

\end{abstract}


\section{Introduction}
\IEEEPARstart{I}{n} the field of autonomous driving \cite{zhou2024drivinggaussian,li2024integrated}, deep learning models \cite{vaswani2017attention} play a pivotal role, largely due to its powerful feature learning capabilities, end-to-end learning processes, and ability to integrate multi-modal data. These strengths contribute to the increased reliability, safety, and efficiency of autonomous driving technologies in real-world scenarios. Driving behavior description \cite{bhattacharyya2022modeling}, a critical sub-task within autonomous driving, involves the deep understanding and precise interpretation of vehicle behavior in various traffic environments. In this task, large language models (LLMs) \cite{2023videochat}, \cite{gao2024vista}, \cite{zhang2023video}, \cite{Maaz2023VideoChatGPT} based on deep learning have demonstrated exceptional performance, particularly due to their strong reasoning and contextual understanding abilities. This makes LLM highly effective for addressing high-level decision-making challenges in autonomous driving. Additionally, LLMs possess the capability to combine driving rules with natural language, enabling them to generate explanatory narratives for driving behaviors.

On the other hand, edge computing \cite{shuvo2022efficient} plays a critical role in the field of autonomous driving, primarily due to its advantages of low latency and rapid response, which meet the real-time and efficient decision-making demands of such systems. In autonomous driving, vehicles may make quick decisions in dynamic and constantly changing traffic environments, such as emergency braking, avoiding pedestrians, or handling complex traffic situations. By offloading computational tasks to edge devices located near the roadside units (RSU) or onboard devices \cite{wu2012cost}, the delays caused by transmitting data to remote cloud servers can be avoided.
Moreover, edge computing reduces bandwidth usage and the burden of data transmission, as it processes and analyzes data locally, sending only essential information to central servers or the cloud. This is particularly important in autonomous driving, where large volumes of sensor data may be continuously processed. The distributed nature of edge computing also enhances system reliability and scalability, preventing the overloading of a single central server and improving overall system stability and responsiveness. Therefore, the deployment of edge computing significantly enhances the real-time performance, safety, and efficiency of autonomous driving systems, especially in handling emergency situations and high-density traffic scenarios.

\IEEEpubidadjcol
In this paper, we propose a framework that integrates large language models (LLMs) with edge devices. This approach combines the strengths of LLMs in understanding complex semantics and reasoning with the powerful image-processing capabilities of visual encoders, resulting in more efficient and flexible descriptions of driving scenes. Specifically, the visual encoder analyzes and extracts key visual features from the driving environment, such as vehicle positions and speed, which are then fed into the LLM for further processing. We deploy the LLMs across multiple roadside units (RSUs), each covering a specific area and interconnected through 5G NR/NSA technology \cite{dogra2020survey}, enabling decentralized deployment. Within this framework, each RSU processes only the traffic data from its coverage area, thereby avoiding redundant operations and mitigating data congestion.

The LLM performs frame-by-frame analysis of driving behavior and road conditions, offering reasoning and explanations, while globally broadcasting warnings about emergencies or hazards, such as overspeeding. To enhance the accuracy of LLMs in processing visual features, we propose a three-stream prompt strategy using multi-modal information, consisting of environmental, agent, and motion streams. These streams convert the extracted features into structured natural language descriptions and reasoning prompts, guiding the LLM to generate context-specific responses. Finally, we introduced a visual access window in the architecture, allowing edge users to interact with the RSUs by uploading traffic information via mobile phones or tablets. This feature effectively addresses the blind spots of fixed collection systems.

The contributions of this letter are summarized as follows:
\begin{itemize}
\item[-] We propose a framework to effectively generate narration and reasoning for driving behavior by integrating LLMs and edge devices.
\item[-] This framework significantly reduces responding time for processing complex driving scenarios on edge, demonstrated a per-frame response time of 0.5 seconds or less.
\item[-] We further propose a multi-modal prompt strategy using environment, agent and motion information. Experimental results indicate that the narration accuracy of all four LLMs exceeded 70\%, with the highest achieving a reasoning accuracy of 81.7\%. \end{itemize}

\section{Task Definition}

Our task focus on describe and interpret driving behaviors in autonomous driving scenarios, with the objective of enabling real-time communication of the LLM-generated descriptions and interpretations between RSUs via 5G NR/NSA. As shown in Figure \ref{fig_1}(a), this task involves deploying an LLM on each RSU, where each RSU functions as an edge server. Any RSU can independently broadcast driving behaviors to the global network, particularly when abnormal or hazardous driving behavior occurs. The RSU that first detects such behavior can quickly communicate with other RSUs. Therefore, this task requires the LLMs deployed on RSUs to effectively describe autonomous driving scenarios and interpret the driving behaviors within those scenarios, while also ensuring a rapid response time for efficient operation.

Our work performs narration and reasoning on each edge device. Narration involves creating coherent natural language descriptions of videos, for example, narration should be able to express details like “a person is crossing the street”. This process requires the model to generate grammatically and semantically fluent text that aligns with human language patterns while accurately reflecting the main content of the video. Reasoning provides a deeper understanding of complex video scenarios, for instance, when a vehicle is stopping in front of the line, the model can infer that the traffic lights turn to red. Reasoning helps link different video segments through logical connections, providing semantic support that makes the descriptions more intelligent.

\section{Method}
\subsection{Overview}
Our task aims to test the performance of LLM on edge devices, including the accuracy of narration and reasoning of driving behavior and the response time.
The framework we designed is LLM deployed on RSUs individually, this setup avoids redundant computations and reduces queuing delays during data transmission, minimizing response latency. Due to RSUs are stationary and may have blind spots. It is necessary to manually upload data from blind spot environments to enable the LLM to consider the entire scene more comprehensively.

In this work, we propose a framework to integrate LLMs with edge devices. We deploy LLMs on each edge device and these edge devices communicate with each other using 5G
NR/NSA technology. 
We also design a visual Q\&A window for real-time interaction. Road users can upload dynamic road conditions to avoid blind spots on RSUs, then consult related information using this window. We further propose a multi-modal prompt strategy to enhance the performance of LLMs.
We collect environment, agent and motion information, and input them into LLMs for reasoning. One effective selection strategy is used to select useful multi-modal information.

\subsection{LLMs with Edge Device}
We propose a framework to integrate LLMs with edge device in this section. In our framework, we use three laptops to simulate three edge servers and a 5G router provided by a network operator for 5G cellular communication. We connected the three RSUs to the 5G router and deployed LLMs on each. A single video clip was split into three sequential parts and fed into the RSUs in order. Our goal is for the LLMs on each RSU to provide coherent reasoning and explanations based on the video content, which will be evaluated against a baseline of human annotations. Additionally, we designed a visualization window at the end of the LLM to simulate road condition information uploaded by pedestrians.

\begin{figure*}[!t]
\centering
\includegraphics[width=5in]{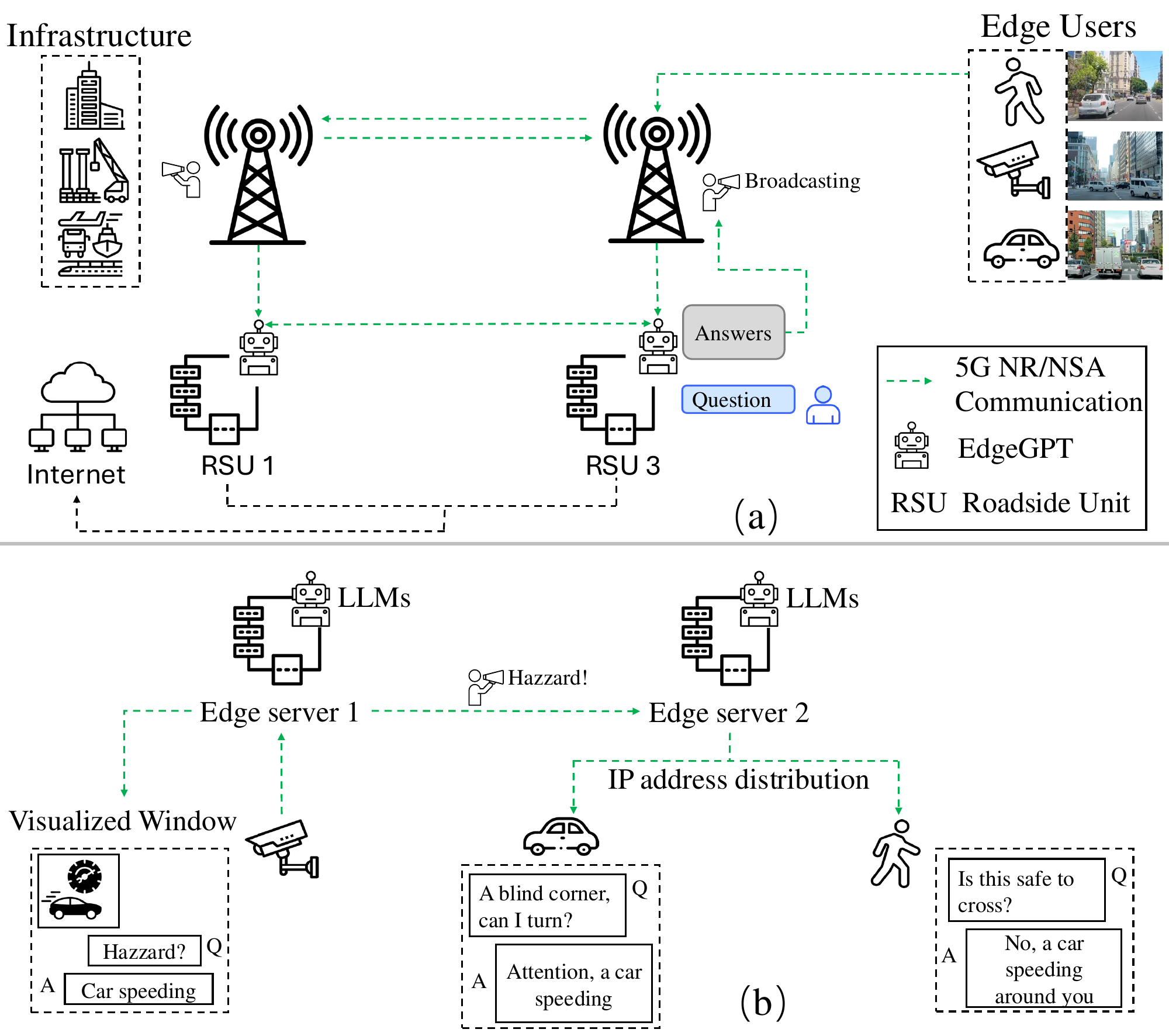}
\caption{(a) Overall system framework. The LLM deployed on RSUs as edge servers, receives input data from edge users via 5G NR/NSA communication technology. It analyzes input data to generate corresponding driving behavior narration and reasoning. Different RSUs can communicate with each other and share information. Finally, the generated textual descriptions are broadcast globally between edge devices. (b) Workflow of our framework deployed on RSU. First, edge users collect surrounding road information and upload to the RSU server using an IP address generated by our framework. LLM generates text-based outputs, which can be accessed through a real-time visualization window for backend queries.}
\label{fig_1}
\end{figure*}

Specifically, the RSUs collects data from vehicles, pedestrians, and infrastructure such as CCTV cameras. Vehicles use onboard sensors for real-time road condition updates, while pedestrians contribute localized data via mobile devices. Data is uploaded to nearby RSUs using dynamically generated IP addresses, and LLMs analyze data to generate text descriptions of road conditions, driving behaviors, and anomalies. Figure \ref{fig_1} (b) shows the workflow of our framework for edge computing, with RSUs processing data locally to reduce latency and meet real-time demands. Processed outputs are visualized for monitoring traffic, driving behavior, and safety.

One key strength of this system is its ability to enable real-time collaboration between RSUs. When an LLM on one RSU detects an event, such as speeding or an accident, this information is immediately communicated to neighboring RSUs. For instance, if the first RSU detects a speeding vehicle, it can alert the second RSU, which can then warn nearby vehicles and pedestrians. This inter-unit communication helps create a safer driving environment by predicting potential dangers before they escalate. The warnings and information are transmitted over the C-V2X network, ensuring timely and reliable message delivery.

\subsection{Multi-modal Prompt Strategy}

The LLMs deployed on the RSUs are not specifically optimized for autonomous driving and driving behaviour tasks. To further enhance the LLM's narration and reasoning capabilities, we design a multi-modal prompt strategy to improve the performance of LLM in narration and reasoning. The prompt includes environment, agent and motion information.

Environment information contains external conditions around the vehicle, such as weather and lighting, which significantly impact driving behavior and decisions. 
It ensures the model considers changes in visibility, road friction, and other environmental factors that may affect vehicle dynamics or necessitate more cautious driving. 
For example, during heavy rain, the model adjusts its reasoning to account for longer stopping distances and reduced lane visibility, enhancing the accuracy of its driving behavior descriptions. 

Agent information directs the model's attention to other entities in the driving environment, such as nearby vehicles and pedestrians. It aids in detecting interactions like lane merging, overtaking, or pedestrian crossings—crucial aspects of safe driving that are challenging to identify without specific prompts. By activating this stream, the model can recognize the presence and behavior of other road users, enabling detailed and context-aware driving behavior descriptions. 

Motion information analyzes the movements of the surrounding vehicles to identify events like sudden braking, sharp turns, and speeding. It enables the model to assess these situations and generate responses, such as recommending slowing down when a vehicle ahead brakes suddenly. This stream detects behaviors that may indicate potential accidents or hazards, providing real-time reasoning for adaptive driving.

We selected 23 environment information to serve as prompts, with each information representing a specific environmental condition, such as rainy day, fog, or nighttime. We first input videos containing one or more of these environmental information, along with their corresponding textual descriptions into the LLM, allowing it to learn and retain these information and their related descriptions. Similarly, we selected 15 agent information as prompts, including pedestrians, vehicles, and cyclists, and 47 motion information, such as turning, stopping, lane changing, and crossing.
After the LLM forms a memory of individual information, we further enrich its understanding of complex scenarios by pairing environmental, agent, and motion information in combinations of two. Using the same strategy, we input these combined scenarios, along with their associated textual descriptions, then fed into the LLM to enhance its memory. This iterative and layered approach ultimately forms a comprehensive multimodal prompts, enabling the LLM to describe and interpret complex scenes from the perspectives of environment, agents, and motion. We use the memory-enhanced LLM as the final version and deploy it on edge devices.

\begin{figure}[!t]
\centering
\includegraphics[width=3.5in]{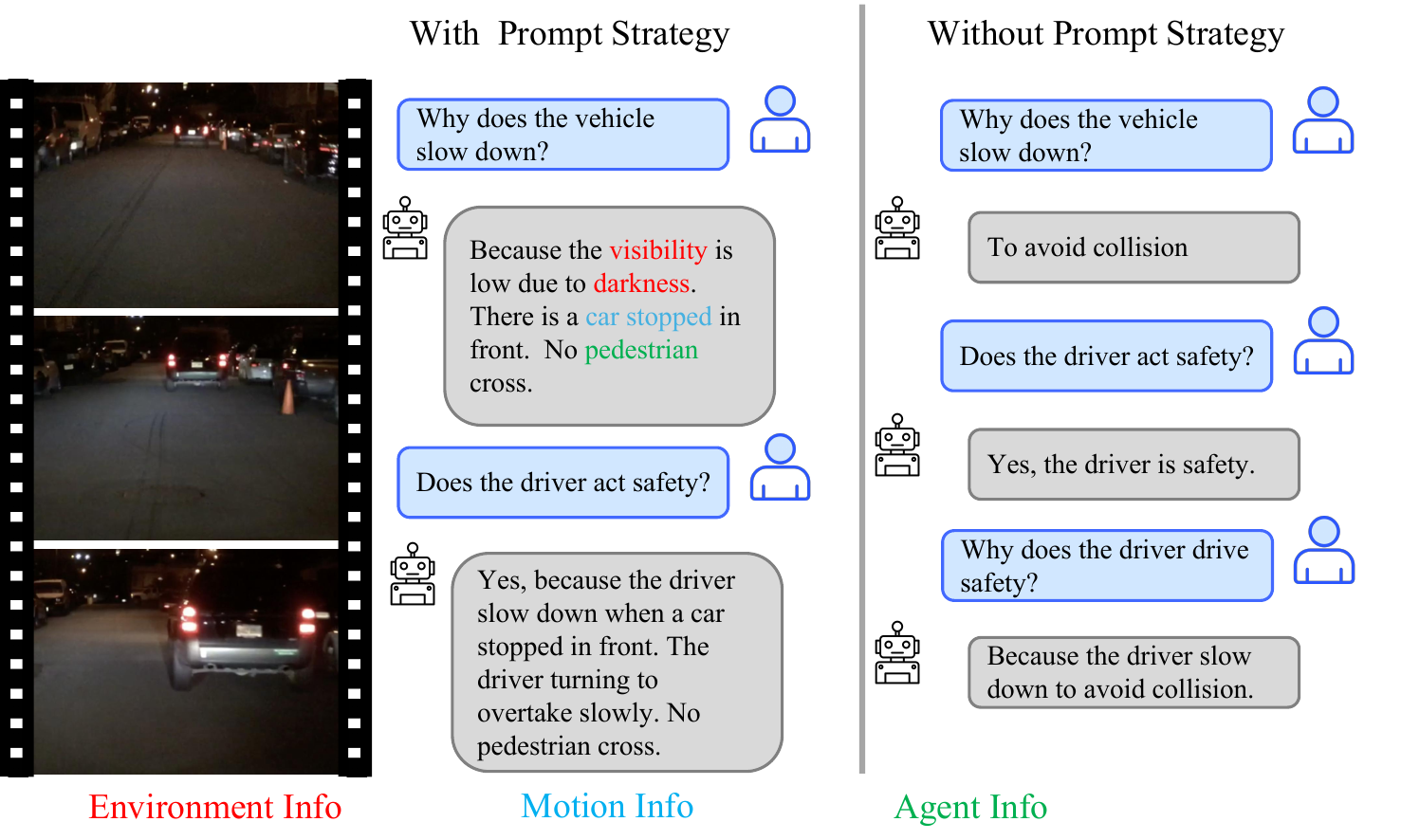}
\caption{Comparison between enabling and disabling three-stream prompt. LLM is able to generate description of driving behivour based on three streams. The observed results can trigger various keywords, such as the environment keyword "visibility," the agent keyword "pedestrian crossing," and the motion keyword "stop," among others.}
\label{fig_2}
\end{figure}
As shown in Figure \ref{fig_2}, when the prompt strategy is enabled, the LLM is able to generate more specific, context-aware outputs that better capture the complexities of driving scenarios. For example, the model can describe a situation where visibility is reduced, a vehicle ahead is prohibited from passing, and no pedestrians are crossing, while also providing an analysis of the underlying driving behaviors.

\section{Experiment and Discussions}
\subsection{Dataset}
The OpenDV-YouTube dataset \cite{yang2024genad}, developed by OpenDriveLab, is a large-scale dataset specifically designed for video-language research in the context of autonomous driving. It includes a wide range of driving videos sourced from YouTube, paired with corresponding text descriptions. These descriptions provide rich annotations that explain driving behaviors, road conditions, and objects encountered in the video. The dataset is intended to support tasks like driving behavior prediction, and autonomous driving research.

\subsection{Experimental Settings}

We implement all models in PyTorch and validated using three NVIDIA RTX 3090 GPUs, each equipped with 24 GB of LPDDR4X memory. It is worth noting that we specifically chose the OpenDV-Youtube dataset for this study, in part because none of the LLMs under evaluation had been previously trained on this dataset. This choice allowed us to fairly and impartially assess the performance of these models on data they had not encountered before, thus offering a more realistic measure of their generalization abilities when applied to new, unseen driving scenarios.

We evaluate model performance by comparing keywords in LLM outputs with the human annotations in the dataset. For instance, if the LLM's response is “3 pedestrians, 1 vehicle in rainy weather”, while the human annotation in the dataset states “3 pedestrians, 1 cyclist, 1 vehicle in rainy weather”, the LLM correctly identifies 3 keywords but misses one, resulting in a narration accuracy of 75\%.
When the LLM explains the scene using phrases like “the motion occurred because of the environment and agent information,” We compare the LLM's reasoning with the human annotations in the dataset, examining the frequency and overlap of keywords to determine the accuracy of the LLM's responses. It is worth noting that if the LLM's reasoning contains structural errors, such as “The weather change is caused by the low speed of vehicles,” it will be considered entirely incorrect due to the flawed cause-effect relationship, even if all the keywords are accurate.

\begin{table}[!t]
\caption{The effect of proposed multi-modal prompt strategy in narration and reasoning task, PS refers to prompt strategy.\label{tab:table1}}

\centering
\renewcommand\arraystretch{1.3}
\tabcolsep=0.1cm

\begin{tabular}{cc|cccc}
\toprule[1.2pt]
\multirow{2}{*}{Task} & \multirow{2}{*}{PS} & \multicolumn{4}{c}{Models} \\
\cline{3-6}
                      &                      & Video Chat  & LLaMA-Ada  & Video-LLaMA  & Video-ChatGPT  \\
\hline
\multirow{2}{*}{Nar.}    & $\checkmark$                   & 76.9\% & 70.3\% & 74.1\% & 78.2\% \\
                      &$\times$                 & 67.2\% & 56.3\% & 59.5\% & 64.9\% \\
\hline
\multirow{2}{*}{Rea.}    & $\checkmark$                    & 71.3\% & 68.1\% & 65.2\% & 81.7\% \\
                      & $\times$                   & 51.4\% & 39.38\% & 44.7\% & 54.5\%\\
\bottomrule[1.2pt]

\end{tabular}
\end{table}

\subsection{Advantage of Multi-modal Prompt Strategy}
We conducted a comprehensive series of tests on Video Chat \cite{2023videochat}, LLaMA Adapter \cite{gao2024vista}, Video LLaMA \cite{zhang2023video}, and Video-ChatGPT \cite{Maaz2023VideoChatGPT} to thoroughly assess their performance in terms of narration accuracy and reasoning correctness, as summarized in Table \ref{tab:table1}. For consistency, the input provided to all four large language models was kept uniform, utilizing raw video data as the primary source. This consistency in input ensured that the results were directly comparable across different models. We designed two distinct sets of experiments: one set with the Prompt Strategy enabled and the other without it. This approach allowed us to effectively evaluate the impact of the Prompt Strategy on the performance of these models, providing insights into how prompt-based optimization influences both narration and reasoning tasks.

We show the result of models with and without prompt strategy in Table \ref{tab:table1}, it is evident that enabling the prompt strategy significantly improves the accuracy of narration and reasoning for all models. Video-ChatGPT performs the best with the prompt strategy enabled, achieving a narration accuracy of 78.2\%  and a reasoning correctness of 81.7\%. However, when the prompt strategy is disabled, these values drop to 64.9\% and 54.5\%, respectively. LLaMA Adapter and Video LLaMA, also show noticeable performance declines when the prompt strategy is disabled, especially in reasoning correctness, where LLaMA Adapter drops from 68.1\% to 39.38\% and Video LLaMA from 65.2\% to 44.7\%. This highlights the crucial role the prompt strategy plays in enhancing the models' understanding and handling of driving behaviors, particularly in reasoning. Overall, the activation of the prompt strategy is essential for improving the performance of all models.

\subsection{Response Speed of LLMs}
We tested the response speeds of these four LLMs, alongside the traditional deep learning method ADAPT \cite{jin2023adapt}. For this aspect of the evaluation, we introduced a variety of conditions by dividing image frames into intervals of 1, 15, and 30. This was done to measure response times under different levels of input frequency. We locally reproduced the ADAPT model and trained it using the OpenDV-Youtube dataset to establish a performance baseline. The results for ADAPT presented in Table \ref{tab:table2} and Table \ref{tab:table3} were obtained from the validation set, providing a benchmark for comparison against the LLMs.

The result in Table \ref{tab:table2} shows that LLMs have significantly fast response times when processing different numbers of image frames. Video-ChatGPT performs the fastest under all conditions, taking 0.3 seconds, 4.4 seconds, and 8.5 seconds to process 1, 15, and 30 frames, respectively. Video Chat, LLaMA Adapter, and Video LLAMA also demonstrate quick response times, particularly when handling smaller batches of images. 
We also implement the conventional deep learning model ADAPT, which shows much longer response times, taking as long as 1173 seconds to process 30 frames, which is significantly slower than the other models. 
This is because that ADAPT is deployed in the local, and has no optimization for the parallel computing. This is also demonstrate the advantage of LLMs in edge environments.

\begin{table}[!t]
\caption{Responding time with different image frames. \label{tab:table2}}
\centering
\renewcommand\arraystretch{1.3}
\tabcolsep=0.15cm
\begin{tabular}{c|ccc}
\toprule[1.2pt]
    Methods & Responding \#1&Responding \#15&Responding \#30\\
\hline
ADAPT &41s   & 619s & 1173s\\
Video Chat&0.4s&6s  & 11.5s \\
LLaMA Adapter &0.4s&5.5s  & 11s\\
Video LLAMA &0.5s&7s   & 13.5s \\
Video-ChatGPT &0.3s &4.4s & 8.5s\\
\bottomrule[1.2pt]
\end{tabular}
\end{table}

\subsection{Discussion}

In this subsection, we discuss extra optimizations in our experiments. When optimizing multi-modal prompts for the LLM, our initial setup involved adding text descriptions for all 30 frames of the video, covering environment, agent, and motion information. After obtaining experimental results from table \ref{tab:table2} for single-frame response speeds, we observed that using only keyframes to add text descriptions for the environment information resulted in the same narration and reasoning accuracy. For agent and motion information, using keyframes along with the trajectory from the preceding and following eight frames allowed the LLM to achieve the results shown in Table \ref{tab:table1}. This optimization significantly reduced the time required for data pre-processing and lowered computational overhead. Additionally, we have listed the performance of the fully trained  deep learning model ADAPT in terms of narration and reasoning in Table \ref{tab:table3}. However, due to the high response speed requirements of this task, the performance of the ADAPT can only serve as a reference.

\section{Conclusion}
In this paper, we proposed framework that integrates large language models and edge devices, along with a multi-modal prompt strategy to enhance the accuracy of narration and reasoning of driving behavior on edge devices. After enabling the multi-modal prompt strategy, the overall performance of the LLM improved significantly. Furthermore, deploying the LLM directly on RSUs via 5G communication technology allows for real-time data processing at the source, significantly reducing the latency associated with data queuing and processing delays. By handling data closer to the source, this approach minimizes waiting times and enhances overall system efficiency.

\begin{table}[!t]
\caption{Accuracy of LLMs and Conventional method ADAPT. LLMs are not trained on the dataset and show remarkable generalization ability. \label{tab:table3}.}
\centering
\renewcommand\arraystretch{1.3}
\tabcolsep=0.1cm
\begin{tabular}{c|ccccc}
\toprule[1.2pt]
 Task & ADAPT &Video-GPT & Video-LLaMA& Video-Chat&LLaMA-Ada \\
\hline

Nar. &89.7\%&78.2 \%&74.1\%  & 76.9\% & 70.3\%\\
Rea. &90.3\%&81.7\%   & 65.2\% & 71.3\%&68.1 \%\\

\bottomrule[1.2pt]
\end{tabular}
\end{table}

\bibliographystyle{ieeetr}
\bibliography{Main/Main}

\begin{thebibliography}{10}

\bibitem{zhou2024drivinggaussian}
X.~Zhou, Z.~Lin, X.~Shan, Y.~Wang, D.~Sun, and M.-H. Yang, ``Drivinggaussian: Composite gaussian splatting for surrounding dynamic autonomous driving scenes,'' in {\em Proceedings of the IEEE/CVF Conference on Computer Vision and Pattern Recognition}, pp.~21634--21643, 2024.

\bibitem{li2024integrated}
X.~Li, X.~Gong, Y.-H. Chen, J.~Huang, and Z.~Zhong, ``Integrated path planning-control design for autonomous vehicles in intelligent transportation systems: A neural-activation approach,'' {\em IEEE Transactions on Intelligent Transportation Systems}, 2024.

\bibitem{vaswani2017attention}
A.~Vaswani, N.~Shazeer, N.~Parmar, J.~Uszkoreit, L.~Jones, A.~N. Gomez, {\L}.~Kaiser, and I.~Polosukhin, ``Attention is all you need,'' {\em Advances in neural information processing systems}, vol.~30, 2017.

\bibitem{bhattacharyya2022modeling}
R.~Bhattacharyya, B.~Wulfe, D.~J. Phillips, A.~Kuefler, J.~Morton, R.~Senanayake, and M.~J. Kochenderfer, ``Modeling human driving behavior through generative adversarial imitation learning,'' {\em IEEE Transactions on Intelligent Transportation Systems}, vol.~24, no.~3, pp.~2874--2887, 2022.

\bibitem{2023videochat}
K.~Li, Y.~He, Y.~Wang, Y.~Li, W.~Wang, P.~Luo, Y.~Wang, L.~Wang, and Y.~Qiao, ``Videochat: Chat-centric video understanding,'' {\em arXiv preprint arXiv:2305.06355}, 2023.

\bibitem{gao2024vista}
S.~Gao, J.~Yang, L.~Chen, K.~Chitta, Y.~Qiu, A.~Geiger, J.~Zhang, and H.~Li, ``Vista: A generalizable driving world model with high fidelity and versatile controllability,'' {\em arXiv preprint arXiv:2405.17398}, 2024.

\bibitem{zhang2023video}
H.~Zhang, X.~Li, and L.~Bing, ``Video-llama: An instruction-tuned audio-visual language model for video understanding,'' {\em arXiv preprint arXiv:2306.02858}, 2023.

\bibitem{Maaz2023VideoChatGPT}
M.~Maaz, H.~Rasheed, S.~Khan, and F.~S. Khan, ``Video-chatgpt: Towards detailed video understanding via large vision and language models,'' in {\em Proceedings of the 62nd Annual Meeting of the Association for Computational Linguistics (ACL 2024)}, 2024.

\bibitem{shuvo2022efficient}
M.~M.~H. Shuvo, S.~K. Islam, J.~Cheng, and B.~I. Morshed, ``Efficient acceleration of deep learning inference on resource-constrained edge devices: A review,'' {\em Proceedings of the IEEE}, vol.~111, no.~1, pp.~42--91, 2022.

\bibitem{wu2012cost}
T.-J. Wu, W.~Liao, and C.-J. Chang, ``A cost-effective strategy for road-side unit placement in vehicular networks,'' {\em IEEE Transactions on Communications}, vol.~60, no.~8, pp.~2295--2303, 2012.

\bibitem{dogra2020survey}
A.~Dogra, R.~K. Jha, and S.~Jain, ``A survey on beyond 5g network with the advent of 6g: Architecture and emerging technologies,'' {\em IEEE access}, vol.~9, pp.~67512--67547, 2020.

\bibitem{yang2024genad}
J.~Yang, S.~Gao, Y.~Qiu, L.~Chen, T.~Li, B.~Dai, K.~Chitta, P.~Wu, J.~Zeng, P.~Luo, J.~Zhang, A.~Geiger, Y.~Qiao, and H.~Li, ``Generalized predictive model for autonomous driving,'' 2024.

\bibitem{jin2023adapt}
B.~Jin, X.~Liu, Y.~Zheng, P.~Li, H.~Zhao, T.~Zhang, Y.~Zheng, G.~Zhou, and J.~Liu, ``Adapt: Action-aware driving caption transformer,'' in {\em 2023 IEEE International Conference on Robotics and Automation (ICRA)}, pp.~7554--7561, IEEE, 2023.

\end{thebibliography}

\end{document}